%% file: main.tex
\documentclass[sigconf]{acmart}
\AtBeginDocument{%
  }

\setcopyright{acmlicensed} 
\copyrightyear{XXXX}
\acmYear{XXXX}
\acmDOI{XXXXXXX.XXXXXXX}

\acmConference[Arxiv Preprint]{Make sure to enter the correct
  conference title from your rights confirmation email}{XXXX XX--XX,
  XXXX}{XXXXXXXX, XX}

\acmISBN{978-1-4503-XXXX-X/XXXX/XX}

\usepackage{multirow}
\usepackage{subcaption}
\usepackage{graphicx}
\usepackage{caption}
\usepackage{subcaption}

\begin{document}

\title{HiMoE: Heterogeneity-Informed Mixture-of-Experts for Fair Spatial-Temporal Forecasting}


\author{Shaohan Yu}
\email{doublepi@buaa.edu.cn}
\affiliation{%
  \institution{Beihang University}
  \city{Beijing}
  \country{China}}

\author{Pan Deng}
\authornote{Corresponding Author}
\email{pandeng@buaa.edu.cn}
\affiliation{%
  \institution{Beihang University}
  \city{Beijing}
  \country{China}}

\author{Yu Zhao}
\email{iyzhao@buaa.edu.cn}
\affiliation{%
  \institution{Beihang University}
  \city{Beijing}
  \country{China}}

\author{Junting Liu}
\email{liujunting@buaa.edu.cn}
\affiliation{%
  \institution{Beihang University}
  \city{Beijing}
  \country{China}}

\author{Zi'ang Wang}
\email{wangziang@buaa.edu.cn}
\affiliation{%
  \institution{Beihang University}
  \city{Beijing}
  \country{China}}
\renewcommand{\shortauthors}{Shaohan Yu et al.}

\begin{abstract}
\input{abstract}
\end{abstract}



\received{20 February 2007}
\received[revised]{12 March 2009}
\received[accepted]{5 June 2009}

\maketitle

\section{Introduction}
\input{introduction}

\section{Preliminary}
\input{preliminaries}

\section{Methodology}
\input{methodology}

\section{Experiment}
\input{experiment}

\section{Related Work}
\input{related-work}

\section{Conclusion}
\input{conclusion}



\bibliographystyle{ACM-Reference-Format}
\bibliography{sample-base}


\end{document}

%% file: abstract.tex
Achieving both accurate and consistent predictive performance across spatial nodes is crucial for ensuring the validity and reliability of outcomes in fair spatial-temporal forecasting tasks. However, existing training methods treat heterogeneous nodes with a fully averaged perspective, resulting in inherently biased prediction targets. Balancing accuracy and consistency is particularly challenging due to the multi-objective nature of spatial-temporal forecasting. To address this issue, we propose a novel Heterogeneity-Informed Mixture-of-Experts (HiMoE) framework that delivers both uniform and precise spatial-temporal predictions. From a model architecture perspective, we design the Heterogeneity-Informed Graph Convolutional Network (HiGCN) to address trend heterogeneity, and we introduce the Node-wise Mixture-of-Experts (NMoE) module to handle cardinality heterogeneity across nodes. From an evaluation perspective, we propose STFairBench, a benchmark that handles fairness in spatial-temporal prediction from both training and evaluation stages. Extensive experiments on four real-world datasets demonstrate that HiMoE achieves state-of-the-art performance, outperforming the best baseline by at least 9.22\% across all evaluation metrics. 

%% file: introduction.tex
\begin{figure}
  \centering
  \includegraphics[width=0.9\linewidth]{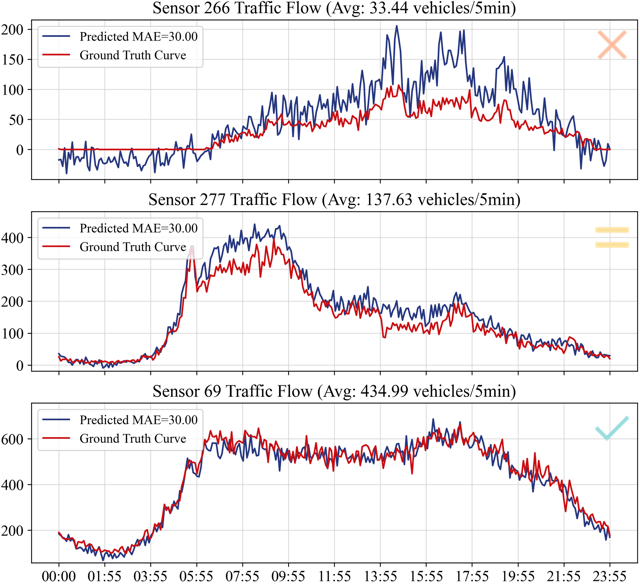}
  \caption{In the PEMS04 dataset, the prediction performances of different nodes under identical Mean Absolute Error (MAE) exhibit substantial disparities, primarily driven by inherent variations in node-specific traffic volumes.}
  \label{fig:intro}
\end{figure}

Spatial-temporal forecasting is a complex multi-objective prediction task, widely applied in fields such as transportation~\cite{dmstgcn,gman,stg-ncde,stnscm,t-mgcn,cchmm}, climate~\cite{pm2.5GNN,GC-LSTM,bggru,mastergnn,ssh-gnn}, and other domains. Ensuring high prediction accuracy at each spatial node is a prerequisite for reliable and usable data predictions. However, existing models evaluate performance solely based on overall performance, overlooking inconsistency in prediction accuracy between different nodes~\cite{fairfor}. In traffic prediction systems, poor prediction performance at specific nodes can lead to missed detection of traffic congestion, which may result in localized traffic paralysis. Achieving both accuracy and uniformity as objectives thus becomes a core challenge in spatial-temporal forecasting tasks. 

However, accurately predicting nodes with heterogeneous temporal trends remains a significant challenge. As shown in Figure~\ref{fig:intro}, different nodes follow distinct daily patterns: Sensor 226 is more active in the evening, Sensor 277 peaks in the morning, while Sensor 69 maintains high activity throughout the daytime. This kind of trend heterogeneity is a fundamental property of spatial-temporal data. This suggests that physical proximity alone is not sufficient for modeling spatial relationships, as it overlooks temporal similarities between nodes. Therefore, modeling inter-node relationships from multiple perspectives is crucial for improving the effectiveness of spatial-temporal forecasting.

In addition, achieving consistent prediction performance across nodes with heterogeneous cardinality remains challenging. As shown in Figure~\ref{fig:intro}, commonly used metrics like Mean Absolute Error (MAE) can be unfair under cardinality heterogeneity. The same MAE leads to a larger relative error for low-value nodes and a smaller one for high-value nodes. Some works try to measure consistency using the variance of node-level MAE, but this introduces bias. Reweighting methods have been proposed to improve training, yet they still rely on these biased evaluation metrics~\cite{fairstg}. Ratio-based metrics like Mean Absolute Percentage Error (MAPE) offer a different view, but they struggle when values are close to zero. Therefore, designing fair ratio-based evaluation methods that can handle cardinality heterogeneity remains an open challenge in spatial-temporal prediction.

Based on the previous discussion, the challenges posed by spatial-temporal data heterogeneity in forecasting tasks can be summarized as follows: (1) How to model inter-node relationships from multiple perspectives to effectively handle trend heterogeneity across nodes with diverse temporal patterns. (2) How to design fair evaluation metrics that are robust to cardinality heterogeneity and reflect prediction quality across nodes with varying magnitudes. (3) How to jointly address trend and cardinality heterogeneity, balancing the dual objectives of accuracy and consistency in spatial-temporal forecasting.

In response to the above challenges, this paper proposes the following methods: (1) We introduce Heterogeneity-Informed Graph Convolutional Network (HiGCN) as the core of the expert model, which leverages multi-graph fusion and edge-level gating to flexibly capture dependencies among trend-heterogeneous nodes. (2) We design STFairBench, a comprehensive benchmark for fair spatial-temporal forecasting, tailored to cardinality heterogeneity. STFairBench enables a more equitable evaluation of prediction performance, improving the uniformity of results across nodes. (3) We present Node-wise Mixture-of-Experts (NMoE), a fine-grained MoE framework designed to address both cardinality and trend heterogeneity. Specifically, NMoE assigns different experts to nodes with varying magnitudes to mitigate performance imbalance caused by cardinality heterogeneity, thereby improving prediction uniformity. Meanwhile, each expert leverages its built-in HiGCN to capture diverse temporal patterns across nodes, thereby enhancing prediction accuracy under trend heterogeneity. This design allows the model to effectively balance the trade-off between accuracy and consistency in spatial-temporal forecasting.

To the best of our knowledge, HiMoE is the first method to identify node-level bias in spatial-temporal predictions arising from existing training strategies and propose corresponding correction methods. Our contributions are as follows:
\begin{itemize}
    \item We propose HiGCN as an integral component of experts and NMoE to manage experts, collaboratively enhancing HiMoE’s ability to make fair predictions.
    \item We propose STFairBench, a benchmark that evaluates node-level prediction ratio errors by incorporating heterogeneous data distribution as an inductive bias. Additionally, we design a tailored training loss within STFairBench to guide the model towards fairer predictions across nodes.
    \item Extensive experiments conducted on four urban computing datasets show that HiMoE consistently achieves state-of-the-art performance, significantly enhancing predictive accuracy and fairness across diverse scenarios, with improvements ranging from 9.22\% to 60.01\% across all metrics.
\end{itemize}

%% file: preliminaries.tex
At each time \( t \), spatial-temporal data is represented as \( X_t \in \mathbb{R}^N \), where \( N \) is the total number of nodes. We denote spatial-temporal data from the time step \( m \) to the time step \( n \) across all nodes as \( \mathbf{X}_{m:n} = \{X_m, ..., X_n\}\in \mathbb{R}^{N \times (n-m+1)} \). The spatial-temporal graph at time \( t \) is expressed as \( \mathcal{G}_t = (V, E, A) \), where \( V \) is the set of \( N \) nodes, \( E \) is the set of edges connecting nodes within \( V \), and \( A \) is the adjacency matrix at time \( t \). Given a graph \( \mathcal{G}_t \) and spatial-temporal data from the previous \( T \) time steps, the goal is to learn a mapping function \( f \) to forecast spatial-temporal data for the next \( T' \) time steps. This mapping can be formally defined as: 

\begin{equation}
\mathbf{\hat{X}}_{t+1:t+T'}=f(\mathbf{X}_{t-T+1:t}, \mathcal{G}).
\label{eq:predict-task}
\end{equation}

%% file: methodology.tex
\begin{figure*}[t]
\centering
\includegraphics[width=0.9\textwidth]{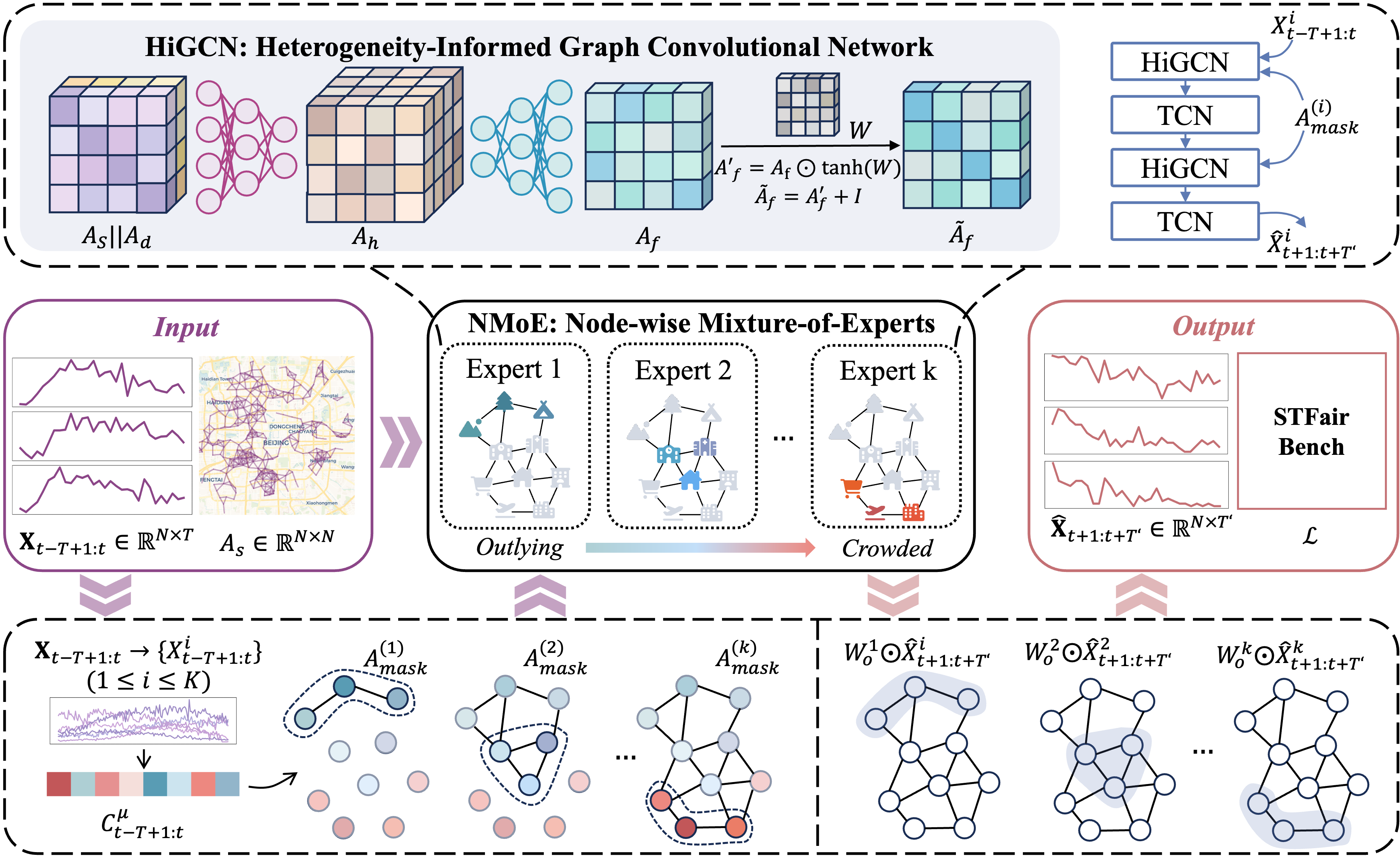}
\caption{The overall architecture of the HiMoE model. The solid boxes represent the model pipeline, while the dashed boxes illustrate how the Mixture of Experts operates and outline the structure of the expert models.}
\label{fig:methodology}
\end{figure*}

This section provides a detailed description of the HiMoE framework, as illustrated in Figure \ref{fig:methodology}. To address the challenges of trend heterogeneity and cardinality heterogeneity in fair spatial-temporal prediction, our model incorporates two dedicated components: a Heterogeneity-Informed Graph Convolutional Network (HiGCN) to handle trend heterogeneity, and a Node-wise Mixture-of-Experts (NMoE) to address cardinality heterogeneity. In addition, we propose STFairBench, a fairness-oriented training and evaluation framework designed to guide the model toward fair spatial-temporal predictions.

\subsection{HiGCN: Heterogeneity-Informed Graph Convolutional Network}
We design HiGCN to address trend heterogeneity. This requires modeling not only the distance-based relationships between nodes but also their trend similarities and dissimilarities. Specifically, we first construct a static adjacency matrix $A_s$ using a Gaussian kernel function to capture the distance-based correlations among nodes:
\begin{equation}
A_{ij} = 
\begin{cases}
\exp\left(-\frac{d_{ij}}{2\sigma^2}\right) & i\neq j,d_{ij} \leq \epsilon, \\
0 & \text{otherwise},
\end{cases}
\label{eq:static-adj}
\end{equation}
where $d_{ij}$ denotes the Euclidean distance between nodes $i$ and $j$, $\sigma$ is a hyperparameter that controls the scale of the Gaussian kernel, and $\epsilon$ is a hyperparameter that sets the maximum allowable distance. A smaller distance between two nodes results in a larger value of $A_{ij}$, indicating a stronger distance-based correlation.

To construct a trend-based correlation adjacency matrix (as illustrated in Figure~\ref{fig:intro}, where different nodes exhibit distinct trends), we define a dynamic adjacency matrix $A_d$ to capture the similarity between the time series of each node:
\begin{equation}
A_{d} = \text{SoftMax}(E^T_{t-T+1:t} E_{t-T+1:t} - \text{diag}(E^T_{t-T+1:t} E_{t-T+1:t})),
\label{eq:dynamic-adj}
\end{equation}
where $E_{t-T+1:t}$ denotes the $\ell_2$-normalized embedding of each node's input time series, and the $\text{diag}(\cdot)$ function extracts the diagonal elements of a matrix. Before applying the softmax function, we remove the diagonal elements of the spatial-temporal adjacency matrix to emphasize the similarities between different nodes.

Based on the constructed static graph $A_s$ and dynamic graph $A_d$, we employ a multi-layer perceptron (MLP) to model the fused relationships between the static and dynamic graphs. Additionally, a tanh-based gating layer is introduced to enable the model to filter out spurious correlations and effectively capture both positive and negative correlations.
\begin{equation}
A_{f} = \text{MLP}(A_d || A_s)\odot \tanh(W_a),
\label{eq:fusion-adj}
\end{equation}
where $(A_d || A_s) \in \mathbb{R}^{N \times N \times 2}$ denotes the concatenation of the dynamic and static adjacency matrices, $\odot$ represents the Hadamard product, and $W_a$ is a learnable parameter matrix. This matrix is normalized to the range $(-1, 1)$ via the $\tanh(\cdot)$ function, allowing the model to capture positive, negative, and neutral correlations between nodes.

Finally, we adopt a graph convolutional network (GCN) based on spectral graph convolution with first-order Chebyshev polynomial approximation to enable information propagation between nodes:
\begin{equation}
h^{(l+1)}=\sigma(\tilde{D}^{-1/2}\tilde{A}_f\tilde{D}^{-1/2}h^{(l)}W^{(l)}),
\label{eq:gcn}
\end{equation}
where $l$ denotes the layer index, $h^{(l)}$ is the input feature at layer $l$, $h^{(l+1)}$ is the output feature at layer $l$, and $\tilde{A} = A + I$ is the adjacency matrix with added self-loops.

To enable the GCN to capture node-level trend correlations, we first obtain the trend-based dynamic adjacency matrix using Equation~\ref{eq:dynamic-adj}. Next, we fuse the static and dynamic graphs via Equation~\ref{eq:fusion-adj}, where spurious correlations are filtered and both positive and negative correlations are enhanced. Finally, the frequency-domain graph convolution defined in Equation~\ref{eq:gcn} is applied to build a node propagation mechanism based on the fused graph. These three steps collectively empower the GCN to effectively handle trend heterogeneity.

\subsection{NMoE: Node-wise Mixture-of-Experts}

As shown in Figure~\ref{fig:intro}, nodes with heterogeneous cardinalities exhibit different prediction performance under the same MAE. This indicates that nodes with varying cardinalities have fundamentally different sensitivities to prediction errors. To address this issue, we introduce the NMoE architecture tailored to cardinality heterogeneity. By assigning nodes with different cardinalities to different expert models, NMoE leverages the specialization of experts in handling heterogeneous node cardinalities, ultimately improving the performance of the model's predictions.

In addition, we adopt a fine-grained sparse Mixture-of-Experts (MoE) framework, which has been widely used in large models. This allows us to increase the model capacity for spatial-temporal prediction while leveraging the parallelism of MoE architectures to accelerate both training and inference. 

The NMoE architecture consists of three components: node cardinality representation learning, a node-level gating mechanism, and the expert model architecture. The node-level gating mechanism uses the learned node cardinality representations to assign different spatial-temporal data to distinct experts for specialized spatial-temporal prediction tasks. The outputs from the various experts are then aggregated via an output gating mechanism. As illustrated in Figure~\ref{fig:methodology}, each expert model comprises four layers combining HiGCN and Temporal Convolutional Network (TCN). In the following sections, we will focus on detailing the node cardinality representation learning and the node-level gating mechanism.

\subsubsection{Node Cardinality Representation Learning}

To obtain a meaningful representation of each node’s cardinality, we consider not only the node’s own statistical features but also the features of its neighbors. This allows us to capture the relative cardinality of each node within its local context.

Given the input data $X_{t-T+1,t}$, we first compute the node-wise mean feature as $C^\mu_{t-T+1,t} = \text{mean}(\textbf{X}_{t-T+1,t})$. We then apply a single-layer spectral graph convolutional network to propagate information among nodes and obtain a local fused representation:
\begin{equation}
\mathcal{R}=\sigma(\tilde{D}^{-1/2}\tilde{A}_s\tilde{D}^{-1/2}C^\mu_{t-T+1,t}W_{r}),
\label{eq:representation-learning}
\end{equation}
where $A_s$ is the static adjacency matrix defined in Equation~\ref{eq:static-adj}, and $W_r$ is a learnable weight matrix. To enhance the robustness of node cardinality representation learning, we apply z-score normalization to the fused representation, yielding the standardized embedding $\hat{\mathcal{R}} = (\mathcal{R} - \mu_\mathcal{R}) / \sigma_\mathcal{R}$. This normalized representation captures the overall cardinality differences across nodes while preserving the relative magnitudes within each local neighborhood.

\subsubsection{Node-wise Gating Mechanism}
The normalized embedding obtained from the representation learning module is fed into the node-wise gating network to enable dynamic assignment of nodes to expert models. This design encourages nodes with similar cardinalities to be handled by the same expert, while nodes with significantly different cardinalities are modeled by different experts.

To this end, we introduce a set of learnable expert center vectors $e = (e_1, \ldots, e_k)$, where $k$ denotes the number of experts. Each expert center corresponds to a different level of node cardinality, with higher-indexed experts representing higher cardinalities. For each node, the probability of being assigned to each expert is computed based on the Euclidean distance between its embedding and the expert centers, enabling an expert selection strategy driven by spatial pattern differences.

Specifically, the distance between node $i$'s representation and expert $j$'s center is computed as $D_{i,j} = \|\hat{\mathcal{R}}_i - e_j\|^2$. The assignment probability $P_{i,j}$, which indicates how likely node $i$ is to be assigned to expert $j$, is then calculated by applying a temperature-controlled softmax function over all experts:
\begin{equation}
P_{i,j} = \frac{\exp\left(-\frac{1}{\tau} D_{i,j}\right)}{\sum_{m=1}^k \exp\left(-\frac{1}{\tau} D_{i,m}\right)}.
\label{eq:expert-prob}
\end{equation}
Here, $\tau$ is a temperature hyperparameter that controls the sharpness of the distribution. Applying sparsity over the expert dimension helps maintain the selectivity of the MoE framework, thereby encouraging different experts to specialize in modeling cardinality heterogeneity nodes.

Based on the node-to-expert assignment probabilities $P \in \mathbb{R}^{N \times k}$, we design both input and output gating mechanisms. The \emph{input gating mechanism} feeds the same spatial-temporal sequence into distinct expert models, each of which performs independent modeling based on its specifically constructed spatial graph. The \emph{output gating mechanism}, guided by a set of learnable parameters, adopts a sparse MoE strategy to autonomously determine and aggregate the outputs from different experts.

\paragraph{Input Gating Mechanism} We design adjacency matrix masks based on the assignment probabilities $P_{i,j}$. By removing certain inter-node connections, each expert is encouraged to focus on spatial nodes with similar cardinality characteristics. As illustrated in Figure~\ref{fig:intro}, nodes with lower cardinality (e.g., Sensor 266) tend to be more sensitive to prediction errors. Therefore, expert models should avoid modeling dependencies between low-cardinality nodes and those with significantly higher cardinalities. To enforce this, we compute a mask to prune specific edges from the fused adjacency matrix defined in Equation~\ref{eq:fusion-adj}:
\begin{equation}
A_{\text{mask}}^{(i)} = P_{1:N,1:i} \times P_{1:N,1:i}^T \, ,
\label{eq:expert-mask}
\end{equation}
where $P_{1:N,1:i}$ denotes the submatrix of $P$ containing all rows and the first $i$ columns. Since each element $P_{i,j} \in (0,1)$, all values in $A_{\text{mask}}^{(i)}$ also lie within the range $(0,1)$.

\paragraph{Output Gating Mechanism} We adopt a sparse MoE strategy consistent with the input gating. A fully learnable gating network is used to adaptively weight and aggregate the outputs from different experts:
\begin{equation}
\hat{\mathbf{X}}_{t+1:t+T'} = \text{sum}((\hat{\mathbf{X}}_{t+1:t+T'}^1 \parallel \dots \parallel \hat{\mathbf{X}}_{t+1:t+T'}^k) \odot \sigma(W_o)),
\label{eq:output-gating}
\end{equation}
where $(\hat{\mathbf{X}}_{t+1:t+T'}^1 \parallel \dots \parallel \hat{\mathbf{X}}_{t+1:t+T'}^k) \in \mathbb{R}^{k\times N\times T'}$ denotes the concatenation of outputs from all $k$ expert models. The output gating weights $\sigma(W_o) \in \mathbb{R}^{k\times N}$ are constrained to the range $(0,1)$, where $\sigma(\cdot)$ denotes the sigmoid activation function. The function $\text{sum}(\cdot)$ aggregates the outputs along the expert dimension $k$, producing the final prediction $\hat{\mathbf{X}}_{t+1:t+T'} \in \mathbb{R}^{N\times T'}$.

\subsection{STFairBench: Spatial-Temporal Fair Benchmark}
\label{sec:stfairbench}

Fair spatial-temporal prediction requires considering both the accuracy and consistency of predictions across different nodes. Existing studies, such as STMoE~\cite{stmoe}, typically use the mean and variance of node-level evaluation metrics (e.g., MAE) as evaluation metrics. However, as shown in Figure~\ref{fig:intro}, conventional metrics like MAE are inherently biased and fail to capture the fairness of prediction performance across nodes. Therefore, there is a pressing need in the spatial-temporal forecasting community for an unbiased method to evaluate both the accuracy and consistency of node-level predictions.

Existing evaluation metrics for spatial-temporal forecasting primarily include Mean Absolute Error (MAE), Root Mean Squared Error (RMSE), and Mean Absolute Percentage Error (MAPE). However, MAPE is sensitive to near-zero denominators, making it unsuitable for use as a loss function during model training. Moreover, MAE and RMSE are inherently biased when evaluating nodes with cardinality heterogeneity. To address these limitations, we propose an unbiased, node-level evaluation metric by formulating the problem from the perspective of a weighted MAE (WMAE):
\begin{equation}
\text{WMAE} = \frac{\text{mean}(\mathbf{X})}{\text{mean}(\mathbf{X}^i)}\frac{1}{T'}\sum_{j=1}^{T'} |X_j^i-\hat{X}_j^i|.
\label{eq:wmae}
\end{equation}
Here, we eliminate the inherent bias of MAE by normalizing it with $\text{mean}(\mathbf{X}^i)$, effectively transforming it into a ratio-like error metric. We then multiply the result by $\text{mean}(\mathbf{X})$ to restore the original data scale. Importantly, both $\mathbf{X}^i$ and $\mathbf{X}$ represent the entire time series. To avoid data leakage, we ensure that when WMAE is incorporated as a training loss, its computation excludes any data from the test set.

Based on the previously computed WMAE for each node, we calculate the mean and standard deviation across all nodes to evaluate both overall accuracy and consistency:
\begin{equation}
\text{MWMAE} = \frac{1}{N}\sum_{i=1}^{N}\text{WMAE}_i,
\label{eq:mwmae-swmae}
\end{equation}
\begin{equation}
\text{SWMAE} = \sqrt{\frac{1}{N} \sum_{i=1}^{N} \left(\text{WMAE}_i - \text{MWMAE}\right)^2}.
\label{eq:mwmae-swmae}
\end{equation}
Here, MWMAE (Mean of WMAE) reflects the average performance across all nodes, while SWMAE (Standard deviation of WMAE) captures the disparity in prediction errors among nodes, serving as a measure of fairness.

Based on the proposed MWMAE and SWMAE metrics, we construct a loss function that explicitly guides the model toward fair spatial-temporal prediction:
\begin{equation}
\mathcal{L}(X, \hat{X}) = \mathcal{L}_{\text{MWMAE}}(X, \hat{X}) + \alpha \mathcal{L}_{\text{SWMAE}}(X, \hat{X}),
\label{eq:loss}
\end{equation}
where $\alpha$ is a hyperparameter that controls the trade-off between minimizing the average prediction error and reducing the disparity in errors across different nodes.

%% file: experiment.tex
\subsection{Experimental Setup}

\subsubsection{Data Descriptions}

We conducted experiments on four datasets: PeMS04, KnowAir, Beijing, and Tongzhou. The PeMS04 dataset is derived from the Caltrans Performance Measurement System (PeMS), a widely used public dataset \cite{astgcn}. This study utilizes only the traffic flow data from PeMS04. KnowAir covers a broad geographical area for PM2.5 forecasting, including several heavily polluted regions across China \cite{pm2.5GNN}. The Tongzhou and Beijing datasets, collected by China Mobile, include road segment speed data for Tongzhou district and real-time population data for urban grids in Beijing. In summary, our datasets encompass various types of data collected across multiple spatial scales, demonstrating their diversity and wide coverage. We apply Z-score normalization to normalize the data. After that, we split the datasets into training, validation, and test data using the ratio of 6:2:2. The detailed information for the four datasets is described in Table \ref{tab:datasets}.

\begin{table}[htbp]
\caption{Detailed information of the four datasets.}
\small
\centering
\begin{tabular}{cccc}
\toprule
\textbf{Dataset} & \textbf{Sensors} & \textbf{Time Interval} & \textbf{Time Steps}\\ \midrule
PEMS04 & 307 & 5 min & 16992 \\ 
KnowAir & 184 & 60 min & 11688 \\ 
Beijing & 500 & 5 min & 26784 \\ 
Tongzhou & 645 & 5 min & 25056 \\ \bottomrule
\end{tabular}
\label{tab:datasets}
\end{table}

\subsubsection{Baselines}
We compare HiMoE with the following baselines: (1) \textbf{Graph WaveNet} \cite{graphwavenet} uses an adaptive adjacency matrix to capture spatial dependencies and combines graph convolution with dilated causal convolution for joint modeling of spatial and temporal dependencies. (2) \textbf{ASTGCN} \cite{astgcn} employs a spatial-temporal attention mechanism and convolution module to learn dynamic correlations. (3) \textbf{MTGNN} \cite{mtgnn} applies a graph-based GNN approach to model multivariate time series, handling scenarios with or without a predefined graph structure. (4) \textbf{D2STGNN} \cite{d2stgnn} integrates dynamic graph learning for spatial dependency modeling and uses both diffusion and inherent models to capture hidden time series. (5) \textbf{TESTAM} \cite{testam} introduces a Mixture-of-Experts model with diverse graph architectures to enhance prediction accuracy. (6) \textbf{BigST} \cite{bigst} uses a linearized global spatial convolution network, reducing complexity for large-scale road network learning and message passing. (7) \textbf{STMoE} \cite{stmoe} offers a model-agnostic framework to mitigate performance bias across road segments, ensuring more balanced predictions.

\subsubsection{Evaluation Metrics}
Based on previous studies, we use MAE, RMSE, and MAPE to evaluate accuracy. Additionally, we adopt STFairBench proposed in Section~\ref{sec:stfairbench} for a fair evaluation of both accuracy and consistency.

\begin{table*}[htbp]
\centering
\caption{The performance on PeMS04 Flow, Beijing Population, Tongzhou Speed, and KnowAir PM2.5 is presented, with the best and second-best performances in each dataset and metrics are bolded and underlined.}
\label{tab:main-result}
\begin{tabular}{c|c|cccccccc}
\toprule
\textbf{Dataset} & \textbf{Metric} & GWNet & ASTGCN & MTGNN & D2STGNN & TESTAM & BigST & STMoE & HiMoE \\
\midrule
\multirow{5}{*}{PeMS04 Flow} 
& MAE  & 18.02 & \underline{16.80} & 17.22 & 19.24 & 18.67 & 17.99 & 17.68 & \textbf{12.36} \\
& RMSE & 29.32 & \underline{27.40} & 28.33 & 31.06 & 30.44 & 29.53 & 28.83 & \textbf{18.45} \\
& MAPE & 12.97 & \underline{11.89} & 12.17 & 14.17 & 12.99 & 12.95 & 12.39 & \textbf{10.00} \\
& MWMAE  & 15.12 & \underline{14.12} & 14.40 & 16.17 & 15.56 & 14.65 & 14.83 & \textbf{10.07} \\
& SWMAE & 22.90 & \underline{21.42} & 22.07 & 24.38 & 23.56 & 22.36 & 22.53 & \textbf{12.24} \\
\midrule
\multirow{5}{*}{Beijing Population} 
& MAE  &100.93 &144.41 & \underline{90.02} &153.70 &136.74 &119.93 & 94.64 & \textbf{60.44} \\
& RMSE &349.49 &513.95 &\underline{296.99} &449.57 &368.61 &423.81 &356.39 & \textbf{132.71} \\
& MAPE & 17.83 & 45.19 & \underline{17.10} & 22.87 & 21.51 & 28.29 & 36.35 & \textbf{14.97} \\
& MWMAE  & 80.10 &111.64 & \underline{71.71} &125.27 &117.60 & 95.46 & 75.07 & \textbf{46.02} \\
& SWMAE &370.65 &532.21 &\underline{319.48} &473.56 &434.07 &442.01 &399.39 & \textbf{127.77} \\
\midrule
\multirow{5}{*}{Tongzhou Speed} 
& MAE  &  3.02 &  2.95 &  2.94 &  3.07 &  3.10 &  3.12 &  \underline{2.93} &  \textbf{2.66} \\
& RMSE &  4.70 & \underline{4.54} &  4.59 &  4.72 &  4.74 &  4.75 &  4.58  &  \textbf{3.86} \\
& MAPE &  8.90 &  \underline{8.57} &  \underline{8.57} &  9.05 &  9.06 &  9.26 &  \underline{8.57} & \textbf{7.42} \\
& MWMAE  &  2.88 &  2.81 &  \underline{2.80} &  2.93 &  2.98 &  2.98 &  \underline{2.80} &  \textbf{2.53} \\
& SWMAE &  3.52 &  \underline{3.39} &  3.46 &  3.50 &  3.51 &  3.49 & 3.46 &  \textbf{2.75} \\
\midrule
\multirow{5}{*}{KnowAir PM2.5} 
& MAE  & 16.77 & 18.33 & 16.18 & 19.29 & 23.61 & 19.71 & \underline{16.08} & \textbf{8.64} \\
& RMSE & 27.26 & 30.01 & 26.22 & 31.04 & 38.18 & 31.69 & \underline{25.92} & \textbf{12.67} \\
& MAPE & 39.21 & 45.30 & \underline{38.58} & 48.66 & 60.59 & 48.64 & 39.30 & \textbf{22.54} \\
& MWMAE  & 15.93 & 17.30 & 15.36 & 18.27 & 22.18 & 18.62 & \underline{15.30} & \textbf{8.17} \\
& SWMAE & 21.31 & 23.07 & 20.41 & 23.67 & 27.90 & 24.07 & \underline{20.23} & \textbf{8.96} \\
\bottomrule
\end{tabular}
\end{table*}

\begin{figure*}[t]
  \centering
  \includegraphics[width=\textwidth]{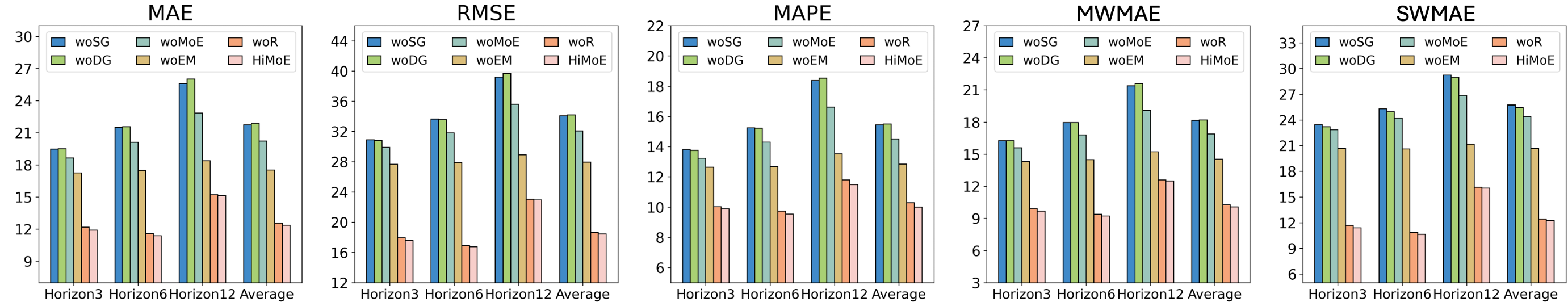}
  \caption{Ablation Study on PeMS04 dataset.}
  \label{fig:ablation-study}
\end{figure*}

\subsubsection{Implementation Details}
We implemented our model and the baseline models using the PyTorch framework on the NVIDIA Tesla V100S GPU. For all datasets, we used the first 12 time steps \(T\) to predict the subsequent 12 time steps \(T'\). The loss function for each baseline model followed its original definition. All models were trained using the Adam optimizer with 500 epochs. Our model was trained with a batch size of 64, a learning rate of 0.003, and a weight decay of 0.0001. For HiGCN, we set the hidden layer dimension of the MLP to 8, the hidden layer dimension of the graph convolution to 64, and used GELU as the activation function.  For NMoE, we set the number of experts to 14 for all datasets. For Equation~\ref{eq:loss}, we set \(\alpha\) to 0.5. For more detailed implementation information, readers can refer to our code repository.

\subsection{Performance Comparision}
According to Table \ref{tab:main-result}, HiMoE outperforms other baselines across datasets with diverse data types and spatial scopes, demonstrating its adaptability to various real-world scenarios. Specifically, on the PeMS04 dataset, our model achieves improvements of 26.43\%, 32.66\%, 15.90\%, 28.68\%, and 43.47\% over the best baseline (ASTGCN~\cite{astgcn}) in MAE, RMSE, MAPE, MWMAE, and SWMAE, respectively. Similarly, on the Beijing dataset, it achieves gains of 32.86\%, 55.31\%, 12.46\%, 35.83\%, and 60.01\%. Additionally, HiMoE delivers improvements of 9.22\%, 14.98\%, 13.42\%, 9.64\%, and 18.88\% on the Tongzhou dataset, and achieves notable progress of 46.27\%, 51.12\%, 41.58\%, 46.60\%, and 55.71\% on the KnowAir dataset.

Compared with models specifically designed for traffic prediction (e.g., GraphWaveNet~\cite{graphwavenet}, D2STGNN~\cite{d2stgnn}, BigST~\cite{bigst}), our model exhibits superior performance on the Beijing and KnowAir datasets, emphasizing HiMoE's scalability across diverse scenarios. Similarly, relative to fully self-learning graph models (e.g., MTGNN~\cite{mtgnn}), our model's results validate that learning dynamic and static graphs is more effective for capturing spatial-temporal dependencies. In contrast to MoE-based approaches (e.g., TESTAM~\cite{testam}), the performance of our model highlights the fine-grained MoE structure and NMoE design. Furthermore, against existing fair prediction models (e.g., STMoE~\cite{stmoe}), our model demonstrates significant advantages in fairness evaluation metrics, showcasing the importance of the synergy between model architecture and training strategy in spatial-temporal prediction tasks. Nevertheless, the final performance of any model is subject to various influencing factors. Therefore, although HiMoE achieves outstanding results as presented in Table \ref{tab:main-result}, these results alone do not comprehensively reflect the effectiveness of each method.

\subsection{Ablation Studies}
In this section, we conduct an ablation study on the PeMS04 dataset to evaluate the effectiveness of each submodule. Five variants of the model are compared: (1) \textit{woSG}: By replacing the fusion adjacency matrix with the dynamic adjacency matrix, the influence of the static adjacency matrix on the model is removed. (2) \textit{woDG}: By substituting the fusion adjacency matrix with a static adjacency matrix, the effect of the dynamic adjacency matrix on the model is eliminated.  (3) \textit{woEM}: The tanh mask applied to the edges in each HiGCN module is removed.  (4) \textit{woMoE}: The MoE structure is removed, leaving only a single expert for predictions. (5) \textit{woR}: Removes the input gating mechanism, distinguishing experts solely via output gating. The prediction results are shown in Figure~\ref{fig:ablation-study}.

\begin{table}[htbp]
\centering
\small
\caption{The overall performance on the PeMS04 Flow dataset is shown, with all models trained using the fairness loss function. The best and second-best performances in each dataset and metric are bolded and underlined.}
\label{tab:lossfunction-study}
\begin{tabular}{c|ccccc}
\toprule
\textbf{Model} &  \textbf{MAE}  & \textbf{RMSE}  & \textbf{MAPE}  & \textbf{MWMAE}  & \textbf{SWMAE}  \\
\midrule
GWNet  & 18.34 & 29.31 & 12.90 & 15.26 & 22.07 \\
ASTGCN         & \underline{16.86} & \underline{26.88} & \underline{12.12} & \underline{14.05} & \underline{20.20} \\
MTGNN          & 17.43 & 28.11 & 12.51 & 14.50 & 21.20 \\
D2STGNN        & 19.48 & 31.04 & 14.14 & 16.24 & 23.55 \\
TESTAM         & 27.06 & 41.40 & 19.42 & 22.55 & 30.85 \\
BigST          & 18.05 & 29.24 & 13.21 & 15.03 & 22.32 \\
STMoE          & 18.00 & 28.86 & 12.73 & 14.96 & 21.72 \\
HiMoE          & \textbf{12.36} & \textbf{18.45} & \textbf{10.00} & \textbf{10.07} & \textbf{12.24} \\
\bottomrule
\end{tabular}
\end{table}

First, replacing the fusion graph with a dynamic or static graph leads to significant performance degradation, demonstrating the necessity of graph learnability. Second, we find that removing the edge mask leads to a decline in prediction accuracy, demonstrating that this mask is indispensable in modeling dependencies. Third, using a single expert model results in a significant performance gap compared to HiMoE, demonstrating that fine-grained MoE greatly enhances predictive capability. Moreover, when the input gating mechanism is removed, the slight performance decline further supports the assumption that reducing the modeling of certain dependencies helps achieve more accurate predictions.

\subsection{Loss Function Studies}

Following the development of the STFairBench, it is crucial to examine whether other models can benefit from its loss function to achieve improved performance. In this section, we apply the loss function introduced in Section~\ref{sec:stfairbench} to train baseline models on the PeMS04 dataset. The performance of the models on PeMS04 is presented in Table~\ref{tab:lossfunction-study}.

\begin{figure}
  \centering
  \includegraphics[width=0.9\linewidth]{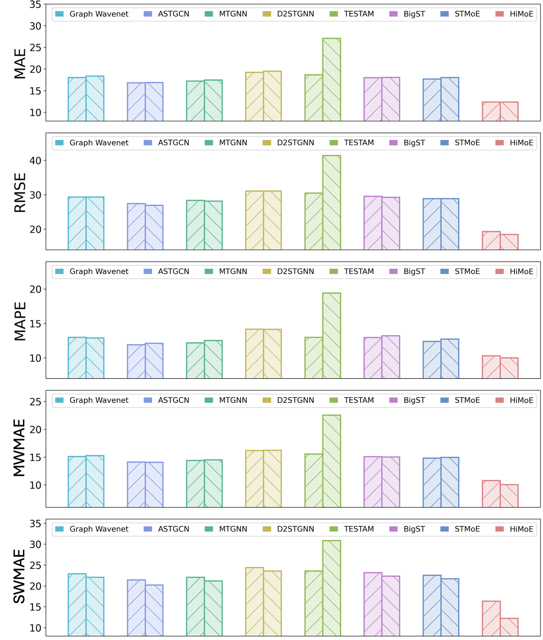}
  \caption{Comparison of baseline model performance with different loss functions: the left bar represents the model performance trained with MAE loss, while the right bar represents the model performance trained with the loss from STFairBench.}
  \label{fig:loss-function}
\end{figure}

First, some models, such as TESTAM, fail to train effectively with the fairness loss function, resulting in performance degradation. Second, for most models (e.g., Graph WaveNet), there is minimal change in MAE, RMSE, MAPE, and MWMAE when using different loss functions. However, SWMAE decreases after training with the fairness loss function. Moreover, HiMoE achieves significant improvements across most evaluation metrics when trained using Equation~\ref{eq:loss} as the loss function compared to using MAE as the loss. Specifically, while there is no difference in MAE, we observed improvements of 4.35\%, 3.00\%, 6.59\%, and 25.06\% in RMSE, MAPE, MWMAE, and SWMAE, respectively. In contrast, the improvements for other models were limited, with maximum gains of -0.36\%, 1.90\%, 0.54\%, 0.50\%, and 5.70\% across these five metrics.

These findings suggest that simply guiding the model through the loss function may not be sufficient to achieve notable improvements in the uniformity of spatial-temporal prediction performance. HiMoE, by incorporating the NMoE design, is better aligned with the fairness loss function, leading to significant improvements in fairness metrics. Additionally, the uniformity evaluation metric strongly correlates with accuracy metrics. Trained with our proposed loss function, HiMoE consistently improves both uniformity and prediction accuracy, demonstrating the effectiveness of the loss function in balancing these objectives.

\begin{figure*}[ht]
  \centering
  \includegraphics[width=0.9\textwidth]{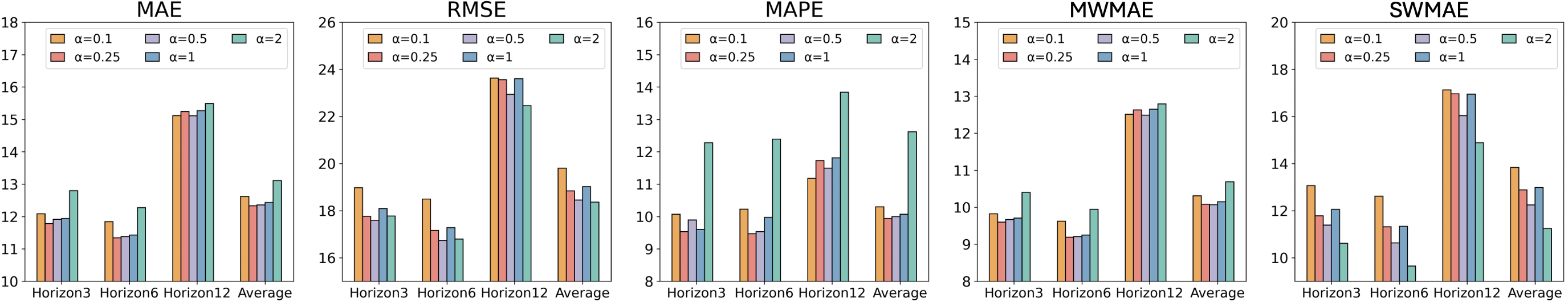}
  \caption{Parameter sensitivity with respect to the loss function coefficient $\alpha$.}
  \label{fig:param-alpha}
\end{figure*}

\begin{figure*}[ht]
  \centering
  \includegraphics[width=0.9\textwidth]{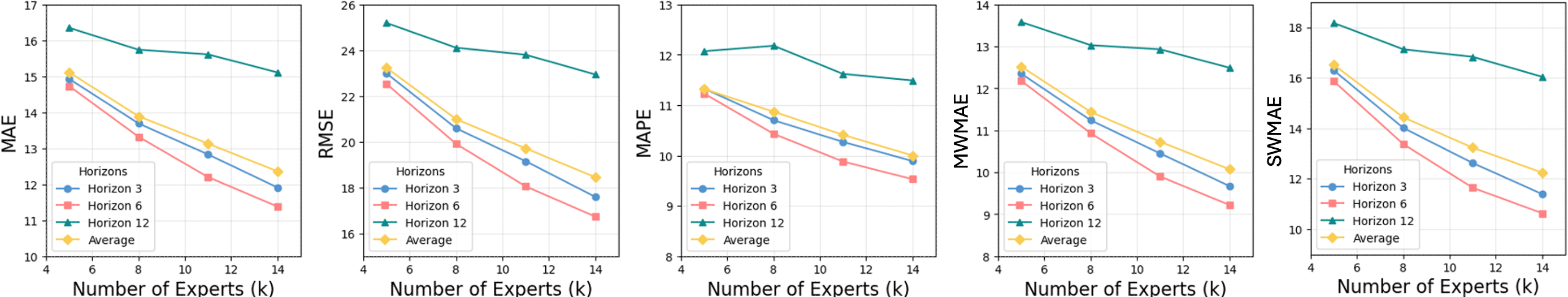}
  \caption{Parameter sensitivity with respect to the number of experts 
$k$.}
  \label{fig:param-k}
\end{figure*}

\subsection{Parameter Sensitiveness}
In this section, we examine the reasonableness of the chosen values for two key hyperparameters: the accuracy-consistency trade-off coefficient $\alpha$ in the loss function, and the number of experts $k$.

The parameter sensitivity of $\alpha$ is shown in Figure~\ref{fig:param-alpha}. It is evident that when $\alpha$ is too large (e.g., $\alpha=2$), the model's accuracy degrades significantly. Conversely, when $\alpha$ is too small (e.g., $\alpha=0.1$), the consistency across experts drops noticeably. Setting $\alpha=0.5$ achieves a better balance across all evaluation metrics.

The parameter sensitivity of the number of experts $k$ is illustrated in Figure~\ref{fig:param-k}. As shown, increasing $k$ consistently improves prediction performance across all metrics. This result supports the hypothesis that spatial-temporal prediction benefits from the parameter scaling law \cite{scaling-law}. Therefore, it is preferable to use a larger number of experts $k$ when hardware resources permit.

\subsection{Efficiency Evaluation}

\begin{table}[htbp]
\small
\centering
\caption{Efficiency evaluation on PeMS04 dataset.}
\label{tab:efficiency}
\begin{tabular}{c|cc}
\toprule
\textbf{Model} &  \textbf{Parameters (k)} & \textbf{Time per Epoch (s)}  \\
\midrule
GWNet    & 311.37k & 129.9s  \\
ASTGCN   & 450.03k & 130.9s \\
MTGNN    & 546.60k & 29.3s \\
D2STGNN  & 12.25k & 101.5s \\
TESTAM   & 212.51k & 83.2s \\
BigST    & 129.58k & 103.8s \\
STMoE    & 1345.02k & 211.2s \\
HiMoE    & 2728.92k & 68.5s \\
\bottomrule
\end{tabular}
\end{table}

In this section, we evaluate the efficiency of the proposed HiMoE model. Specifically, we compare the number of parameters and the training time per epoch on the PeMS04 dataset, as summarized in Table~\ref{tab:efficiency}.

As shown in the table, HiMoE possesses the largest number of parameters among all compared models. Surprisingly, despite its high model complexity, HiMoE achieves the second fastest training time per epoch, outperforming several models with significantly fewer parameters. Compared with other MoE-based spatial-temporal models like STMoE and TESTAM, HiMoE is the first to adopt a fine-grained expert mixture framework. It uses 14 efficient, low-parameter experts, while STMoE employs 4 GWNet-based experts and TESTAM includes 3 temporal modeling experts. The lightweight design of each expert ensures fast computation. In addition, the large number of experts and the parallel nature of MoE further speed up training. 

As computing infrastructure continues to evolve, larger models with more parameters can be better supported and more efficiently executed. HiMoE demonstrates the feasibility of applying fine-grained MoE to spatial-temporal modeling, achieving both faster training and better performance.

\subsection{Case Study}

\begin{figure*}[ht]
  \centering
  \begin{subfigure}[b]{0.32\textwidth}
    \centering
    \includegraphics[height=4cm]{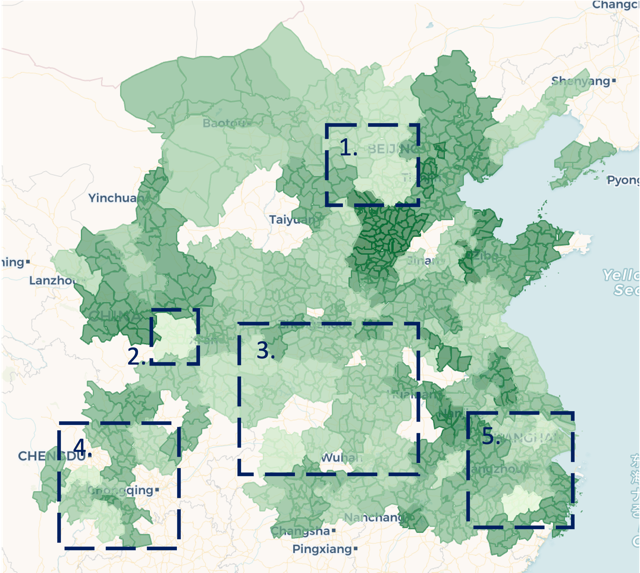}
    \caption{Mean}
    \label{fig:case-mean}
  \end{subfigure}
  \begin{subfigure}[b]{0.32\textwidth}
    \centering
    \includegraphics[height=4cm]{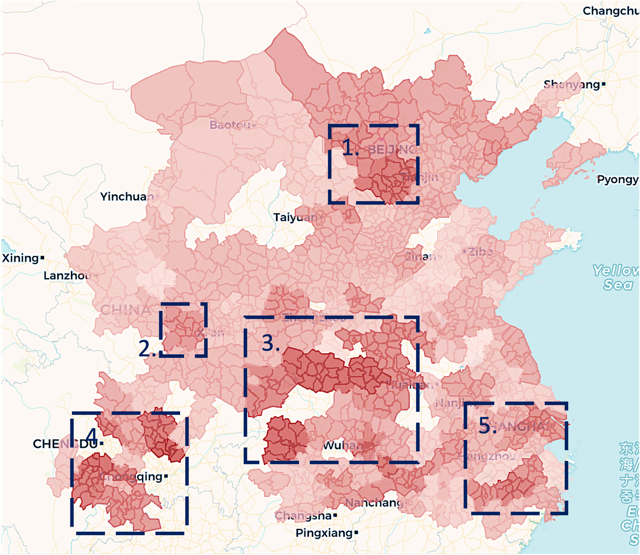}
    \caption{HiMoE + MAE Loss}
    \label{fig:case-mae}
  \end{subfigure}
  \begin{subfigure}[b]{0.32\textwidth}
    \centering
    \includegraphics[height=4cm]{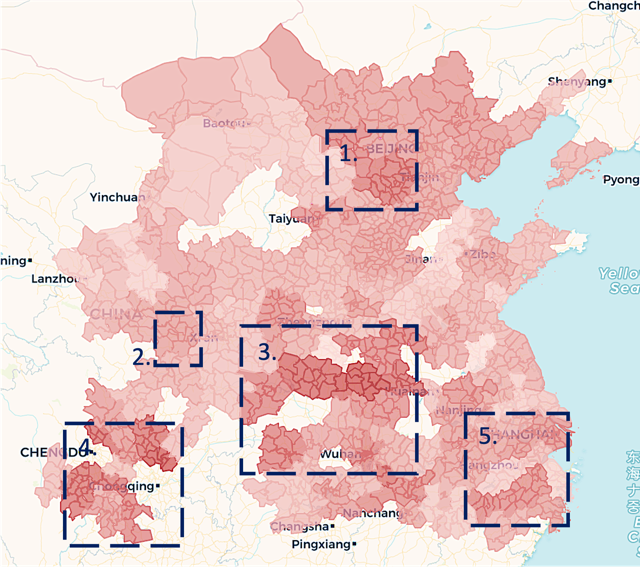}
    \caption{HiMoE + STFairBench Loss}
    \label{fig:case-fair}
  \end{subfigure}
  \caption{Case study on KnowAir dataset.}
  \label{fig:case}
\end{figure*}

In this section, we conduct experiments on the KnowAir dataset to validate the phenomenon illustrated in Figure~\ref{fig:intro}, where nodes with heterogeneous cardinalities exhibit different sensitivities to the same MAE. Figure~\ref{fig:case-mean} visualizes pollution concentrations across different regions in the KnowAir dataset, with darker colors indicating higher concentrations. We highlight five low-concentration regions with dashed boxes—these correspond to nodes with relatively low cardinality.

Figures~\ref{fig:case-mae} and~\ref{fig:case-fair} compare the prediction performance of models trained with different loss functions, where darker colors indicate higher WMAE values. Notably, under the MAE-based loss (Figure~\ref{fig:case-mae}), the five low-cardinality regions exhibit significantly worse prediction accuracy. In contrast, the model trained with the loss function proposed in STFairBench (Figure~\ref{fig:case-fair}) performs better in these regions. Moreover, Figure~\ref{fig:case-fair} displays a more balanced overall color distribution compared to Figure~\ref{fig:case-mae}, indicating higher prediction consistency.

%% file: related-work.tex
\subsection{Spatial-Temporal Data Forecasting}
Spatial-temporal graph neural networks (STGNNs) have proven to be effective in modeling intricate relationships within spatial-temporal data \cite{st4ug-survey}. Traditional models establish robust baseline models by leveraging a variety of temporal and spatial modeling techniques \cite{dmstgcn,stg-ncde,stnscm,stsgcn,agcrn}. In contrast, contemporary approaches focus on harnessing innovative learning methods to fully exploit the untapped potential of existing models \cite{std-mae,megacrn,stssl,himnet,trafficstream}.

Nevertheless, these models rarely recognize the multi-objective nature of spatial-temporal forecasting tasks. They often focus on improving overall prediction accuracy, while paying insufficient attention to the consistency of predictions at the individual node level. Some models recognize the importance of consistency in spatial-temporal forecasting \cite{fairst,tfgnn}. FairFor addresses this issue by generating group-relevant and group-irrelevant representations through adversarial learning \cite{fairfor}. STMoE, on the other hand, leverages spatial-temporal representation learning to capture the unique patterns of each node and employs different experts to handle these distinct patterns, thus ensuring more balanced predictions in spatial-temporal forecasting \cite{stmoe}. FairSTG focuses on achieving fair predictions by optimizing the loss function \cite{fairstg}. It introduces a mechanism that dynamically adjusts the weights of nodes based on their prediction performance, aiming to reduce performance disparities across different nodes. However, existing spatial-temporal fairness methods rely on biased node evaluation metrics, which fail to effectively address the issue of unfairness.

\subsection{Mixture of Experts}
The Mixture of Experts (MoE) model, proposed by Shazeer et al. \cite{moe}, is a framework designed to improve overall performance and efficiency by combining the complementary capabilities of multiple expert models. This architecture dynamically assigns the input data to specialized experts according to the characteristics of the task, allowing the model to achieve exceptional results in complex and diverse tasks \cite{moe-survey}. MoE has seen extensive application in a variety of fields, such as Large Language Model (LLM) \cite{metamoe,openmoe,glam} and image recognition \cite{vmoe,ride,m3vit}. Recently, DeepSeek has demonstrated the potential of fine-grained Mixture-of-Experts (MoE) to develop robust language models, making a substantial impact in the field \cite{deepseek}.

MoE has also been widely applied in spatial-temporal forecasting. Wu et al. \cite{stmoge} utilized MoE to model different types of crime, capturing both unique and shared patterns to enable more robust multi-type crime modeling and support collective crime prediction. Jiang et al. \cite{cpmoe} applied MoE to real-world congestion prediction, enhancing accuracy in dynamic traffic environments by introducing two specialized experts designed to capture stable trends and periodic congestion patterns. Lee et al. \cite{testam} proposes TESTAM, which employs three distinct temporal modeling experts to capture complex spatial-temporal traffic trends. By modeling temporal dynamics from multiple perspectives, TESTAM achieves superior forecasting performance. However, these models often use a limited number of experts, failing to fully leverage the powerful modeling capabilities of fine-grained MoE, resulting in suboptimal performance. Moreover, a limited number of experts in sparse MoE architecture reduces the diversity of available experts, leading to the same expert handling multiple tasks and resulting in inflexible routing. Therefore, applying fine-grained sparse mixture-of-experts (SMoE) to spatial-temporal prediction is urgently needed.

%% file: conclusion.tex
In this work, we identify a previously overlooked fairness issue in evaluating spatial-temporal forecasting performance and propose the HiMoE framework to address this challenge, achieving both accurate and consistent predictions. Specifically, we observe that nodes with heterogeneous cardinality exhibit vastly different prediction performance under the same level of MAE. Because the same MAE means less for high-cardinality nodes and more for low-cardinality ones, high-cardinality nodes seem to perform better, which leads to unfair evaluation.

Cardinality heterogeneity undermines consistency in forecasting, while trend heterogeneity poses challenges to prediction accuracy. To achieve accurate spatial-temporal forecasting, it is essential to model spatial relationships from multiple perspectives. To this end, we propose HiGCN, a Heterogeneity-Informed Graph Convolutional Network that models trend correlations among nodes through a multi-graph fusion mechanism. Furthermore, we propose the Node-wise Mixture-of-Experts (NMoE) module, which allocates different experts to nodes based on their cardinality, effectively promoting performance consistency. Finally, we introduce STFairBench, a comprehensive benchmark that guides the model toward fairness in both training and evaluation, addressing the limitations of traditional metrics.

By addressing the challenges of spatial-temporal heterogeneity across the architecture, training, and evaluation stages, our proposed framework achieves at least a 9.22\% improvement across all evaluation metrics. Moreover, this work is the first to reveal the bias in existing evaluation methods and to demonstrate the effectiveness of fine-grained MoE in the spatial-temporal domain.